\pdfoutput=1

\documentclass[11pt]{article}

\usepackage[preprint]{acl}

\usepackage{times}
\usepackage{latexsym}

\usepackage[T1]{fontenc}

\usepackage[utf8]{inputenc}

\usepackage{microtype}

\usepackage{inconsolata}

\usepackage{graphicx}
\usepackage{booktabs}
\usepackage{multirow}
\usepackage{tcolorbox}
\usepackage{cleveref}
\crefformat{section}{\S#2#1#3}
\crefformat{subsection}{\S#2#1#3}
\crefformat{subsubsection}{\S#2#1#3}
\crefrangeformat{section}{\S#3#1#4 to~\S#5#2#6}
\crefmultiformat{section}{\S#2#1#3}{ and~\S#2#1#3}{, #2#1#3}{ and~#2#1#3}
\Crefformat{figure}{#2Fig.~#1#3}
\Crefmultiformat{figure}{Figs.~#2#1#3}{ and~#2#1#3}{, #2#1#3}{ and~#2#1#3}
\Crefformat{table}{#2Tab.~#1#3}
\Crefmultiformat{table}{Tabs.~#2#1#3}{ and~#2#1#3}{, #2#1#3}{ and~#2#1#3}
\Crefformat{appendix}{#2Appx.~\S#1#3}
\crefformat{algorithm}{Alg.~#2#1#3}
\Crefformat{equation}{#2Eq.~#1#3}

\usepackage[dvipsnames]{xcolor}
\usepackage{listings}
\lstdefinelanguage{json}{
    basicstyle=\small\ttfamily,
    numbers=left,
    numberstyle=\scriptsize,
    breaklines=true,
    frame=lines,
    backgroundcolor=\color{gray!10},
    showstringspaces=false,
    string=[db]{"},
    stringstyle=\color{green!50!black},
    morestring=[s][\color{black}]{\ \ "}{":},
    keywordstyle=\color{blue},
    keywords={true,false,null},
    literate=
     *{0}{{{\color{red}0}}}{1}
      {1}{{{\color{red}1}}}{1}
      {2}{{{\color{red}2}}}{1}
      {3}{{{\color{red}3}}}{1}
      {4}{{{\color{red}4}}}{1}
      {5}{{{\color{red}5}}}{1}
      {6}{{{\color{red}6}}}{1}
      {7}{{{\color{red}7}}}{1}
      {8}{{{\color{red}8}}}{1}
      {9}{{{\color{red}9}}}{1}
      {.}{{{\color{red}.}}}{1}
      {:}{{{\color{gray}{:}}}}{1}
      {,}{{{\color{gray}{,}}}}{1}
      {\{}{{{\color{gray}{\{}}}}{1}
      {\}}{{{\color{gray}{\}}}}}{1}
      {[}{{{\color{gray}{[}}}}{1}
      {]}{{{\color{gray}{]}}}}{1},
}
\usepackage{xspace}
\newcommand{\bench}{\mbox{\textsc{OmniStruct}}\xspace}

\title{\bench: Universal Text-to-Structure Generation across Diverse Schemas}

\author{
James Y. Huang\textsuperscript{\rm 1}~~~
Wenxuan Zhou\textsuperscript{\rm 1}~~~
Nan Xu\textsuperscript{\rm 1}~~~
Fei Wang\textsuperscript{\rm 1}~~~
Qin Liu\textsuperscript{\rm 2}\\
\textbf{Sheng Zhang}\textsuperscript{\rm 3}~~~
\textbf{Hoifung Poon}\textsuperscript{\rm 3}~~~
\textbf{Muhao Chen}\textsuperscript{\rm 2}\\
{\textsuperscript{\rm 1}}University of Southern California~~~
{\textsuperscript{\rm 2}}University of California, Davis~~~
{\textsuperscript{\rm 3}}Microsoft Research\\
\texttt{\{huangjam, zhouwenx, nanx, fwang598\}@usc.edu}\\
\texttt{\{qinli, muhchen\}@ucdavis.edu}~~~
\texttt{\{shezhan, hoifung\}@microsoft.com}\\
  }

\begin{document}
\maketitle

\begin{abstract}
  The ability of Large Language Models (LLMs) to generate structured outputs that follow arbitrary schemas is crucial to a wide range of downstream tasks that require diverse structured representations of results such as information extraction, table generation, and function calling. While modern LLMs excel in generating unstructured responses in natural language, whether this advancement translates to a strong performance on text-to-structure tasks remains unclear. To bridge this gap, we first introduce \bench, a comprehensive benchmark for assessing LLMs' capabilities on diverse text-to-structure tasks such as information extraction, table generation, and function calling. We build \bench by identifying existing datasets across a wide range of tasks that are suitable for a structured answer format, and adapting them under a unified text-to-structure problem setting. To facilitate the development of efficient text-to-structure models, we collect high-quality training data via synthetic task generation. Without using any supervised data for \bench tasks, our experiments demonstrate the possibility of fine-tuning much smaller models on synthetic data into universal structured generation models that can rival the performance of GPT-4o.
\end{abstract}

\section{Introduction}
Rapid advancements in Large Language Models (LLMs) have significantly broadened their application beyond merely generating natural language responses \cite{achiam2023gpt,grattafiori2024llama,yang2024qwen2}. One such application is text-to-structure generation, where a structured format such as JSON is used to precisely represent and organize the task outputs~\citep{openai-structured-outputs}. Many downstream tasks, such as various information extraction tasks \cite{wang2023instructuie}, table generation \cite{wu-etal-2022-text-table}, and automated function calling \cite{patil2024gorilla}, greatly benefit from text-to-structure as it defines a clear and unambiguous answer format and therefore simplifies answer parsing, evaluation, and integration with downstream applications. 

Despite the success of LLMs on natural language generation tasks, realizing universal text-to-structure generation presents unique challenges to LLMs, as it requires a clear understanding of structural requirements and precise organization of results. While some text-to-structure tasks have been studied in prior works \cite{patil2024gorilla,tam-etal-2024-speak,tang-etal-2024-struc}, they generally focus on individual tasks without a unified problem definition. Efforts to unify different text-to-structure tasks have almost exclusively focused on information extraction tasks \cite{lu-etal-2022-unified,wang2023instructuie} such as Named Entity Recognition (NER), Relation Extraction (RE), and Event Extraction (EE). However, the answer schemas designed for IE do not generalize to other tasks with much more diverse answer structures. As a result, they fail to provide a holistic understanding of \textit{Universal Text-to-Structure} capabilities across diverse tasks and schemas.

To bridge this gap, we first introduce \bench, a comprehensive benchmark for evaluating universal text-to-structure generation of LLMs. \bench is a diverse collection of text-to-structure tasks adapted from %
various resources of
information extraction, table generation, and function calling tasks. We construct \bench by first identifying existing tasks where the answer can be non-trivially represented in a structural format. This usually implies that the expected answer contains multiple interconnected components that benefit from a structured organization of the answer. Next, we convert all collected tasks into a unified schema-following text-to-structure task. We specifically adopt JSON as the unified answer format due to its versatility and wide use, making it ideal for representing diverse answer schemas and integrating with downstream pipelines. Each instance consists of a task instruction that states the problem itself in natural language, a format instruction that includes a JSON schema to define the expected answer format, and a JSON ground truth answer that follows the associated JSON schema.

We conduct a comprehensive evaluation of LLMs across different model families and sizes on \bench. While many open-sourced models have been specifically fine-tuned to follow JSON schemas \cite{grattafiori2024llama,yang2024qwen2}, there still exists a significant gap between cost-efficient open-sourced models and state-of-the-art proprietary models like GPT-4o~\citep{hurst2024gpt}. Inspired by the success of model distillation \cite{zhou2024universalner} in developing much smaller yet strong specialized models, we propose to synthesize high-quality text-to-structure supervision data from GPT-4o. Specifically, we start from a large pool of NLP task instructions from diverse sources and prompt GPT-4o to identify tasks suitable for text-to-structure. Then we expand this pool by iteratively proposing new tasks based on existing tasks, and augment new task instructions with model-generated schemas and answers. Without using any supervised data, fine-tuning much smaller models on our distilled supervision data significantly boosts the model's zero-shot text-to-structure capabilities on \bench, rivaling the performance of GPT-4o.

Our contribution is two-fold. First, we present \bench, a comprehensive benchmark for assessing broad text-to-structure capabilities of LLMs under a unified problem definition. Second, we propose a novel pipeline for synthesizing diverse text-to-structure data to facilitate the development of cost-efficient universal text-to-structure models, and demonstrates the effectiveness of our approach.

\section{Related Work}
\subsection{Text-to-Structure}

Several text-to-structure tasks have been studied individually with unique ways of defining the answer format. For example, InstructUIE \cite{wang2023instructuie} proposes multi-task supervised instruction tuning for generative information extraction under a unified task schema. UniversalNER \cite{zhou2024universalner} introduces a multi-turn conversational framework for open named entity recognition. \citet{wu-etal-2022-text-table} propose the task of text-to-table by adapting existing table-to-text datasets. Struc-Bench \cite{tang-etal-2024-struc} extends the text-to-table generation tasks into more diverse formats, including XML and LaTeX. LiveSum \cite{deng-etal-2024-text} explores more challenging cases of text-to-table generation that goes beyond pure text extraction and requires reasoning over extracted information. \citet{tam-etal-2024-speak} investigate the effect of structured format restrictions on reasoning and classification tasks with simple schemas. More recently, function calling \cite{qin2024toolllm,deng-etal-2024-text} emerges as another important text-to-structure application where the function name and arguments need to be precisely organized in a structured format to ensure successful execution. Built upon existing structured generation datasets, \bench presents a comprehensive benchmark for evaluating universal structured generation under a unified problem definition.

Previous work has also studied the broader JSON generation task that is not explicitly tied to any specific downstream task. IFEval \cite{zhou2023instruction} investigates the general instruction-following capability of LLMs by introducing various types of constraints into the prompts, with JSON formatting being one of the constraints included in the benchmark. However, all JSON formatting examples, in addition to being very limited in number, are open-ended generation tasks without any schema requirement. Therefore, the evaluation is purely based on the validity of the JSON object without any verification of schema adherence. FoFo \cite{xia-etal-2024-fofo} extends IFEval to more diverse domains and formats, but their evaluation focuses exclusively on format adherence and does not consider the content quality. JSONSchemaBench \cite{geng2025generating} presents a study focusing purely on schema adherence without any task defined. In other words, the goal is to generate an arbitrary JSON object that follows the given schema, and evaluation is based on schema adherence alone. We believe that none of these works fully reflects the nature of text-to-structure tasks in real-world applications, where the tasks have much clearer objectives with well-defined answer formats and emphasis on both content quality and format adherence. In contrast, \bench encompasses a wide range of text-to-structure tasks rooted in common downstream applications, making it suitable for providing a universal assessment of text-to-structure capabilities.

\subsection{Constrained Decoding}

A closely related line of research to text-to-structure generation is constrained decoding \cite{willard2023efficient,beurer2024guiding,dong2024xgrammar}, which describes a group of decoding algorithms that impose direct constraints on the tokens or their distributions a model can generate at each step. By pruning candidate tokens that violate the format restrictions, constrained decoding can guarantee the validity of the answer even if the model does not fully understand and comply with the format instruction on its own \cite{geng2025generating}. While constrained decoding mostly targets format adherence, \bench assesses both the format and the quality of the answer across a broad range of text-to-structure tasks and schemas. In this work, we focus on evaluating and improving the model’s own capability of completing text-to-structure generation tasks by following task instructions and answer schemas provided in the prompt.

\begin{figure*}[ht]
    \centering
    \includegraphics[width=\textwidth]{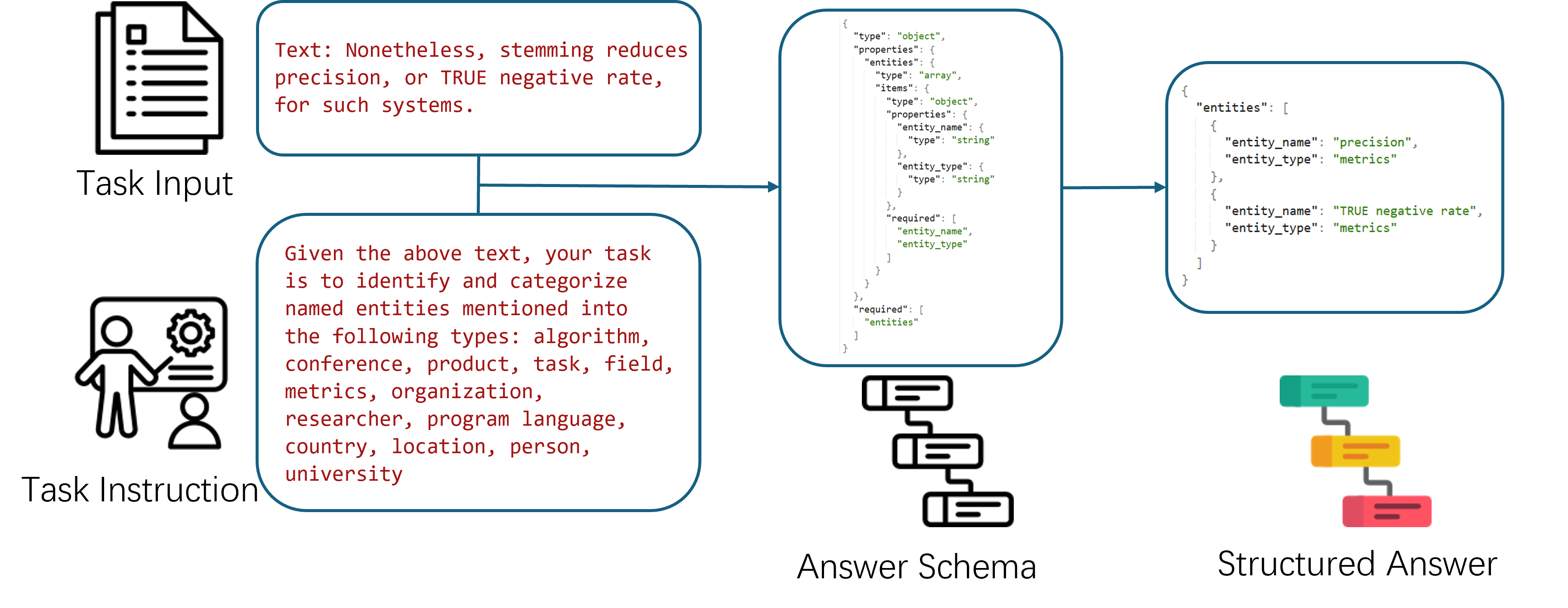}
    \caption{Overview of our unified problem definition for text-to-structure tasks in \bench. Take NER as an example, each task instance consists of a task instruction and input that jointly defines the task, an answer JSON schema that defines the expected answer structure, and a JSON groundtruth answer that strictly follows the given schema.}
    \label{fig:bench_main}
\end{figure*}

\section{\bench}
In this section, we first provide a unified problem definition for text-to-structure tasks in \bench, followed by details about the data collection process, and statistics of each individual dataset included in \bench.

\subsection{Problem Definition}
We first provide a unified problem definition for text-to-structure generation tasks. \Cref{fig:bench_main} shows an overview of how we define a text-to-structure generation task. Among various structural formats, we specifically pick JSON as the universal format for text-to-structure generation due to its versatility and wide use \cite{grattafiori2024llama,yang2024qwen2}. Each text-to-structure task is formulated as schema-following JSON generation, where the goal is to generate the correct task output in JSON format that strictly follows the JSON schema provided by the task prompt. Each instance consists of three components: a task instruction that describes the task at hand (e.g., NER, function calling, etc), a format instruction that contains the JSON schema defining the expected output format, and a ground truth answer written in JSON format. The answer generated by a model is evaluated based on the validity of the JSON answer, its adherence to the given JSON schema, and the correctness of its content. Answers that can not be correctly parsed (either due to being an invalid JSON or missing required fields in the schema) are treated as incorrect regardless of their content.

\begin{tcolorbox}[colback=white, colframe=gray!80!gray, boxrule=1pt, title=Prompt for Universal Text-to-Structure]

\textbf{System Prompt}: You are a helpful assistant. You must write your answer in JSON format according to the given JSON schema. Do not include anything other than the JSON answer in your response.
\newline

Schema: \{answer\_schema\}
\newline

\textbf{User Prompt}: \{task\_instruction\}

\end{tcolorbox}

\subsection{Data Collection}
To build \bench, we start from identifying tasks that are suitable for text-to-structure generation. Specifically, the output of such tasks should be represented in JSON format in a non-trivial way. In principle, any task output can be written in JSON format trivially by converting the original output into a JSON object with a single "output" field. In this work, we are more interested in tasks where the expected output is inherently structured and contains multiple interconnected components that benefit from a structured representation. One such example would be NER, where the output should contain a list of entity names and their corresponding types. For each selected text-to-structure task, we convert the original task ground truth into JSON format and build JSON schemas for the expected output format. Depending on the nature of the task, we either manually write a task-wide JSON schema (e.g., NER, RE) or we automatically construct instance-specific JSON schemas 
based on the ground truth output. In both cases, we make sure that the ground-truth output is always a valid instance under the JSON schema. The resulting benchmark covers a diverse set of tasks ranging from traditional data extraction tasks, including NER, RE, and text-to-table, to new frontiers evaluating modern LLM skills such as function calling and reasoning. Below, we provide details about the collection of text-to-structure tasks in \bench.

~\\\textbf{Named Entity Recognition (NER)}~ aims at identifying and categorizing named entities from a given piece of text. To cover a diverse set of domains and entity types,  we include CoNLL 2003 \cite{tjong-kim-sang-de-meulder-2003-introduction} and all five domain sets of CrossNER \cite{liu2021crossner} spanning AI, literature, music, politics, and science. The task-wide answer schema is defined as a list of dictionaries, with each dictionary consisting of the name of the extracted entity and its corresponding entity type. 

~\\\textbf{Relation Extraction (RE)}~ aims at identifying and categorizing the relation between pairs of entities from a given piece of text. \bench includes ADE corpus \cite{gurulingappa2012development}, CoNLL 2004 \cite{roth-yih-2004-linear}, and SciERC \cite{luan-etal-2018-multi} that covers medical, news, and science domains, respectively. Similar to NER, the task-wide answer schema is defined as a list of dictionaries, with each dictionary consisting of the names of the subject and object entities, and the relation type between them.

~\\\textbf{Text-to-Table}~ transforms selected types of information from a given piece of text into a tabular format. For example, given a brief biography of a person, we may want to extract basic information about the person, including name, birth date, nationality, etc.  \bench includes WikiBio \cite{wu-etal-2022-text-table}, RotoWire \cite{wiseman-etal-2017-challenges}, and LiveSum \cite{deng-etal-2024-text}, sourced from Wikipedia pages and sports commentaries. Since the answer schema varies based on the specific instance, we first convert the original textual answer into JSON format using row and column names as field names, and then automatically construct the JSON schema with the GenSON \footnote{\url{https://github.com/wolverdude/genson/}} library.

~\\\textbf{Function Calling}~ requires LLMs to assemble a function call with all the necessary arguments to solve a task given one or more available functions. It is crucial for LLMs to generate function calls that strictly follow the given function definitions, as slight deviation from predefined schemas may cause the function call to completely fail during execution. For this task, we focus on BFCL \cite{berkeley-function-calling-leaderboard}, which covers a diverse set of problems with varying difficulties, from simple ones that mostly test argument extraction, to challenging ones that involve selecting and planning sequential and parallel function calls. Since we focus on text-to-structure generation rather than automating function execution capabilities, \bench only includes the 400 simple problems from BFCL. The original BFCL data comes with function definitions similar to JSON schemas for each instance, but uses a slightly different set of keywords (e.g., ``parameters'' in place of ``properties'' in a standard JSON schema). Hence, we replace those keywords so that the problem setting is consistent across \bench tasks.

\begin{table}[t]
\small
\centering
\setlength{\tabcolsep}{2pt}
\begin{tabular}{lcccc}
\toprule
\textbf{Task} & \textbf{Dataset} & \textbf{\# test} & \textbf{Avg. \# field}\\ 
\midrule
\multirow{6}{*}{NER} & CrossNER AI & 431 & 3 &\\
& CrossNER literature & 416 & 3\\
& CrossNER music & 465 & 3\\
& CrossNER politics & 650 & 3\\
& CrossNER science & 543 & 3\\
& CoNLL 2003 & 3453 & 3\\
\midrule
\multirow{3}{*}{RE} & ADE & 428 & 4\\
& CoNLL 2004 & 288 & 4\\
& SciERC & 397 & 4 \\
\midrule
\multirow{3}{*}{Text-to-Table} & WikiBio & 500 & 4.1\\
& RotoWire & 500 & 40.4\\
& LiveSum & 754 & 18\\
\midrule
Function Calling & BFCL & 400 & 2.9\\
\midrule
\multirow{2}{*}{Reasoning} & GSM8K & 1318 & 3\\
& Meeting Plan & 1000 & 4\\
\midrule
\end{tabular}
\caption{Details about each individual dataset included in \bench. We can see that \bench covers a diverse set of tasks and schemas.}
\label{tab:stats}
\end{table}

\paragraph{Reasoning} covers a broad range of tasks that involve math, logic, planning, etc., with some of these tasks having a structured answer format. For example, solutions to math problems such as GSM8K \cite{cobbe2021training} can be represented as a combination of the step-by-step reasoning and the final answer. Planning tasks such as Meeting Plan \cite{zheng2024natural} require a structured representation of the meeting schedule. While assessing the reasoning capabilities of LLMs is beyond the scope of \bench, we are still interested in the extent to which structural requirements affect answer quality in reasoning tasks, as recent works have suggested conflicting evidence \cite{tam-etal-2024-speak,geng2025generating}.

\begin{figure}[t]
    \centering
    \includegraphics[width=0.5\textwidth]{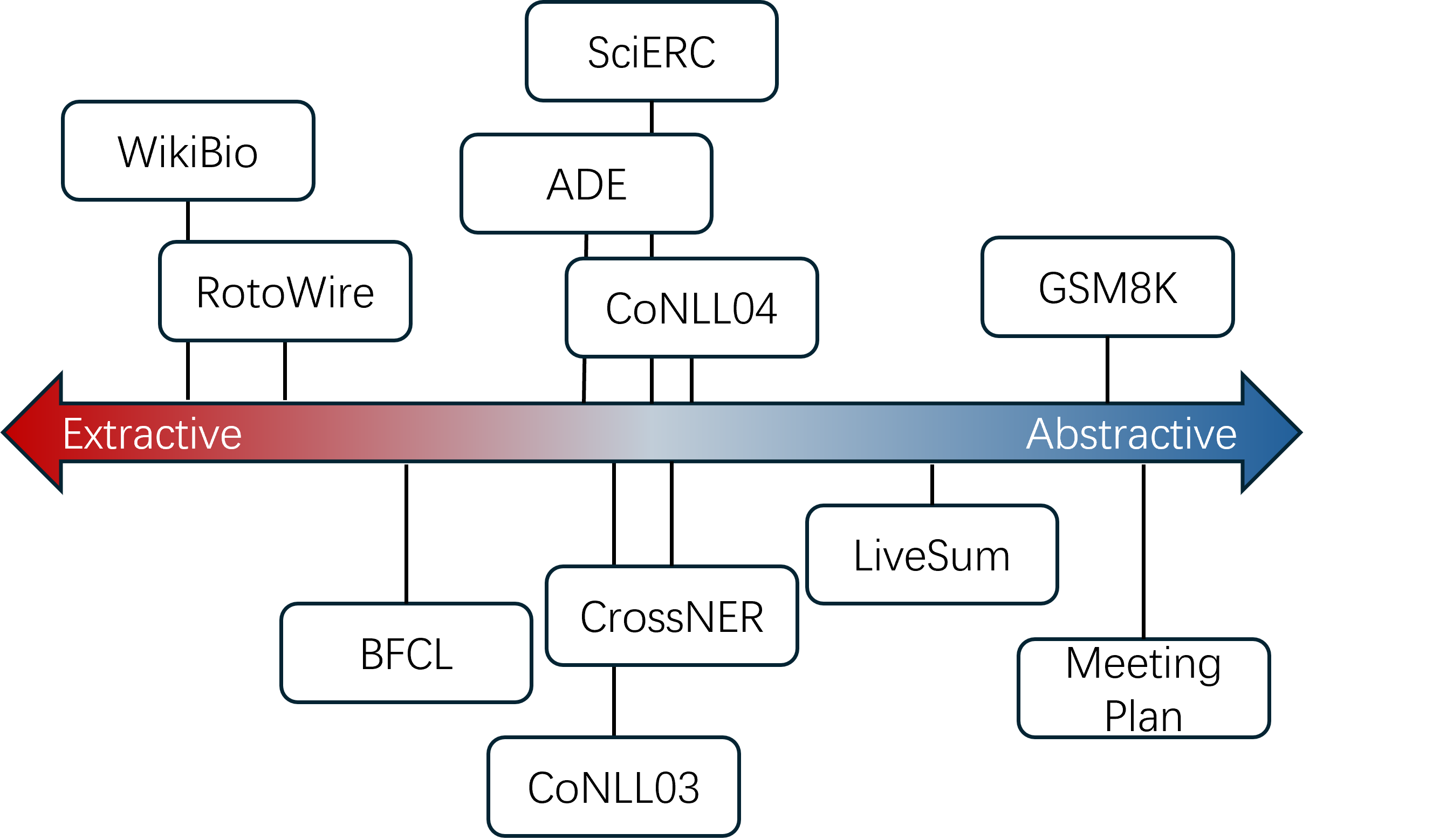}
    \caption{Spectrum of \bench tasks based on how extractive the task is.}
    \label{fig:bench_spectrum}
\end{figure}

\subsection{Statistics}
\label{sec:3.3}
In \Cref{tab:stats}, we provide detailed statistics of \bench. Our benchmark covers a diverse set of schemas with varying number of fields and structural depth. NER and RE tasks tend to have a small number of distinct fields due to their relatively simple schemas, but may have a longer ground-truth answer (as measured by the number of fields in the answer) if multiple entities or relations are extracted. On the other hand, text-to-table and function calling tasks on average have more distinct fields since they have more complicated schemas that specify each different type of information to be extracted. From a different perspective, we can also place these tasks on a spectrum based on how extractive the task is, as shown in \Cref{fig:bench_spectrum}. For example, text-to-table tasks are generally more extractive since the tasks are mostly about finding the correct information. In other words, the JSON answer here is essentially a reorganization of the given information presented in a structured format instead of natural language. LiveSum is a notable exception, since this task is specifically designed to require some reasoning when filling in the table. NER and RE tasks are placed in the middle since they involve both extracting entity names and inferring entity types. On the other end, reasoning tasks like math are more abstractive and require additional skills and complex inference to solve, in which case JSON is mostly used as a means of precisely communicating the final result. Examples of \bench tasks can be found in \Cref{app:bench}.

\begin{table*}[ht]
\small
\centering
\setlength{\tabcolsep}{4pt}
\begin{tabular}{lccccccccc}
\toprule
\multirow{2}{*}{\textbf{Model}} & \multicolumn{6}{c}{\textbf{NER}} & \multicolumn{3}{c}{\textbf{RE}} \\ 
\cmidrule(lr){2-7} \cmidrule(lr){8-10}
& \textbf{AI} & \textbf{literature} & \textbf{music} & \textbf{politics} & \textbf{science} & \textbf{CoNLL03} & \textbf{ADE} & \textbf{CoNLL04} & \textbf{SciERC} \\
\midrule
GPT-4o & 64.4 & 62.5 & 69.6 & 73.9 & \textbf{68.9} & \textbf{73.1} & \textbf{70.3} & 33.0 & 10.8 \\
GPT-4o Structured Outputs & \textbf{64.7} & 60.7 & 68.0 & \textbf{74.0} & 68.6 & 71.5 & 69.8 & \textbf{34.5} & 9.2\\
\midrule
Llama3.1-8B-Instruct & 57.3 & 54.7 & 60.8 & 62.3 & 56.5 & 63.2 & 63.4 & 25.6 & 4.8 \\
Qwen2.5-32B-Instruct & 57.1 & 61.6 & 65.2 & 71.3 & 59.7 & 72.5 & 68.7 & 22.1 & \textbf{11.7}\\
Qwen2.5-14B-Instruct & 58.3 & \textbf{64.5} & 70.6 & 69.8 & 59.7 & 67.4 & 60.2 & 33.5 & 5.3 \\
Qwen2.5-7B-Instruct & 50.9 & 57.0 & 56.4 & 67.7 & 62.1 & 68.4 & 54.3 & 27.9 & 3.9 \\
Phi-4-14B & 61.7 & 61.4 & \textbf{70.7} & 70.9 & 65.6 & 68.8 & 58.5 & 32.7 & 8.4\\
\bottomrule
\end{tabular}

\begin{tabular}{lcccccc}
\toprule
\multirow{2}{*}{\textbf{Model}} & \multicolumn{3}{c}{\textbf{Text-to-Table}} & \textbf{Function Calling} & \multicolumn{2}{c}{\textbf{Reasoning}} \\  
\cmidrule(lr){2-4} \cmidrule(lr){5-5} \cmidrule(lr){6-7}
& \textbf{WikiBio} & \textbf{RotoWire} & \textbf{LiveSum} & \textbf{BFCL} & \textbf{GSM8K} & \textbf{Meeting Plan} \\
\midrule
GPT-4o & \textbf{57.9} & \textbf{87.5} & 57.3 & \textbf{84.3} & 95.2 & \textbf{32.4}\\
GPT-4o Structured Outputs & 57.5 & 87.4 & \textbf{57.8} & 81.0 & \textbf{96.0} & 30.4\\
\midrule
Llama3.1-8B-Instruct & 52.8 & 82.7 & 39.4 & 66.8 & 70.5 & 8.9\\
Qwen2.5-32B-Instruct & 54.4 & 82.0 & 46.8 & 84.0 & 85.1 & 17.1\\
Qwen2.5-14B-Instruct & 52.8 & 86.0 & 41.8 & 80.8 & 91.7 & 17.4\\
Qwen2.5-7B-Instruct & 51.9 & 83.7 & 34.6 & 80.8 & 78.5 & 6.8\\
Phi-4-14B & 51.8 & 82.1 & 43.0 & 78.8 & 94.2 & 0.2\\
\bottomrule
\end{tabular}
\caption{Benchmarking performance of LLMs across different families and sizes on \bench. GPT-4o outperforms other open-sourced models consistently on all studied text-to-structure tasks, while smaller models also perform well on tasks such as NER and RE.}
\label{tab:bench}
\end{table*}

\section{Evaluation}
In this section, we provide details about how we evaluate various LLMs on our proposed \bench, followed by a discussion about the main findings and insights we get from our benchmark evaluation.
\subsection{Experiment Setting}
We evaluate LLMs across different model families and sizes on \bench, including GPT-4o (08-06;~\citealt{achiam2023gpt}), Llama3 (3.1-8B-Instruct;~\citealt{grattafiori2024llama}), Qwen2.5 (7B/14B/32B-Instruct;~\citealt{yang2024qwen2}), and Phi-4 (14B;~\citealt{abdin2024phi}). We also evaluate GPT-4o with the Structured Outputs API \footnote{\url{https://platform.openai.com/docs/guides/structured-outputs}}, which uses constrained decoding techniques to guarantee schema adherence. We set the temperature to 0 for all models. All models are evaluated under a zero-shot setting, where the model is given a system prompt that outlines the format restriction in the form of a JSON schema, and a user prompt that contains the task instruction. For NER and RE tasks, we follow \citet{zhou2024universalner} and report the strict entity-level micro F1 score, which requires an exact match of both entity names and entity/relation types. For text-to-table tasks, we report the average Error Rate across all ground truth cells, where a cell is considered correct if the prediction exactly matches the ground truth. For function calling, we reuse the evaluation pipeline for the original benchmark and report the accuracy based on abstract syntax tree matching. For reasoning tasks, we report the accuracy following their original evaluation protocols.

\subsection{Main Results}
As shown in \Cref{tab:bench}, GPT-4o demonstrates superior overall performance on \bench compared to other smaller open-sourced models. The advantage of GPT-4o is more prominent on challenging tasks such as LiveSum and Meeting Plan, which are considered to be more abstractive as we discussed in \Cref{sec:3.3} and require strong reasoning and planning skills to perform. In addition, we do not observe further improvement from constrained decoding with Structured Outputs API, as we will show in the next section that directly prompting GPT-4o already gives outputs that strictly follow the schema. On the other hand, smaller models have already made impressive progress on many NER and RE datasets, sometimes even outperforming GPT-4o in the zero-shot setting. When comparing models with similar sizes, Qwen2.5-7B-Instruct outperforms Llama3.1-8B-Instruct on function calling, but loses its advantage when evaluated on information extraction tasks. Phi-4 demonstrates very strong performance on NER and RE, but performs very poorly on the meeting planning tasks. Overall, among the open-sourced models we evaluated on \bench, there is no clear winner across all tasks.

\begin{figure}[t]
    \centering
    \includegraphics[width=0.4\textwidth]{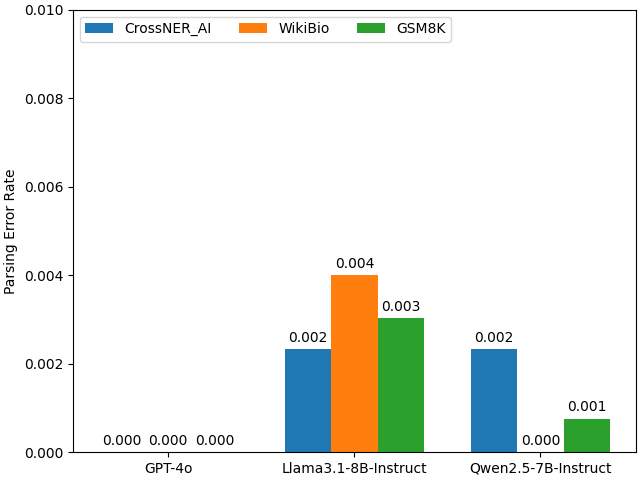}
    \caption{Parsing Error Rate of different LLMs. Parsing errors are rare in general and thus have little impact on the model's performance.}
    \label{fig:bench_parsing_error}
\end{figure}

\subsection{How Well Do LLMs Follow JSON Schemas?} To understand how much performance drop can be attributed to parsing errors, we calculate the number of occurrences in which the generated JSON answer is either an invalid JSON or missing a required field from the given schema. As shown in \Cref{fig:bench_parsing_error}, we find that parsing errors are generally very rare (< 0.5\%). Answers generated by prompting GPT-4o usually don't have any parsing errors, which explains why constrained decoding with Structured Outputs API does not bring further improvement. Weaker models like Llama3.1-8B-Instruct and Qwen2.5-7B-Instruct produce parsing errors only on very rare occasions. This is likely due to the fact that modern LLMs are usually trained on JSON data \cite{grattafiori2024llama,yang2024qwen2} and thus can follow JSON schema very well. However, as shown by the results on \bench, format adherence does not necessitate high-quality answers.

\begin{table*}[ht]
\small
\centering
\begin{tabular}{lccccccccc}
\toprule
\multirow{2}{*}{\textbf{Model}} & \multicolumn{6}{c}{\textbf{NER}} & \multicolumn{3}{c}{\textbf{RE}} \\ 
\cmidrule(lr){2-7} \cmidrule(lr){8-10}
& \textbf{AI} & \textbf{literature} & \textbf{music} & \textbf{politics} & \textbf{science} & \textbf{CoNLL03} & \textbf{ADE} & \textbf{CoNLL04} & \textbf{SciERC} \\
\midrule
GPT-4o & 64.4 & 62.5 & 69.6 & 73.9 & 68.9 & \textbf{73.1} & \textbf{70.3} & 33.0 & \textbf{10.8} \\
Llama3.1-8b-instruct & 57.3 & 54.7 & 60.8 & 62.3 & 56.5 & 63.2 & 63.4 & 25.6 & 4.8 \\
\bench-8B & \textbf{64.7} & \textbf{65.4} & \textbf{71.1} & \textbf{76.9} & \textbf{69.4} & 71.0 & 67.7 & \textbf{33.3} & 8.4 \\
& ({\color{Green}+7.4}) & ({\color{Green}+10.7}) & ({\color{Green}+10.3}) & ({\color{Green}+14.6}) &({\color{Green}+12.9}) & ({\color{Green}+7.8}) & ({\color{Green}+4.3}) & ({\color{Green}+7.7}) & ({\color{Green}+3.6})\\
\bottomrule
\end{tabular}

\begin{tabular}{lcccccc}
\toprule
\multirow{2}{*}{\textbf{Model}} & \multicolumn{3}{c}{\textbf{Text-to-Table}} & \textbf{Function Calling} & \multicolumn{2}{c}{\textbf{Reasoning}} \\  
\cmidrule(lr){2-4} \cmidrule(lr){5-5} \cmidrule(lr){6-7}
& \textbf{WikiBio} & \textbf{RotoWire} & \textbf{LiveSum} & \textbf{BFCL} & \textbf{GSM8K} & \textbf{Meeting Plan} \\
\midrule
GPT-4o & \textbf{57.9} & \textbf{87.5} & \textbf{57.3} & \textbf{84.3} & \textbf{95.2} & \textbf{32.4}\\
Llama3.1-8b-instruct & 52.8 & 82.7 & 39.4 & 66.8 & 70.5 & 8.9\\
\bench-8B & 55.7 & 84.0 & 38.7 & 71.8 & 77.2 & 13.3\\
& ({\color{Green}+2.9}) & ({\color{Green}+1.3}) & ({\color{Red}-0.7}) & ({\color{Green}+5.0}) & ({\color{Green}+6.7}) & ({\color{Green}+4.4})\\
\bottomrule
\end{tabular}

\caption{Benchmarking performance of \bench-8B. We highlight the performance change of \bench-8B compared to its base model Llama3.1-8B-Instruct in brackets. Synthetic instruction tuning data facilitates noticeable capability improvement of the small model on text-to-structure generation, even surpassing GPT-4o on NER and RE.}
\label{tab:sft}
\vspace{-2em}
\end{table*}

\section{Text-to-Structure Instruction Tuning}

\subsection{Data Synthesis}
Given the evaluation results on \bench, we have demonstrated GPT-4o's superior text-to-structure capability across a wide range of tasks. More cost-efficient models, despite some of them being explicitly optimized for JSON generation, still trail behind state-of-the-art large-scale models. While the overall performance gap between GPT-4o and more compact open-sourced models is not easy to fill, previous work \cite{zhou2024universalner} has demonstrated the possibility of distilling powerful LLMs into much smaller, more cost-efficient task-specialized models. By fine-tuning on high-quality model-synthesized data, a small student model can match the performance of a much stronger teacher model in a specialized field, while being significantly more efficient to deploy. Inspired by the success of model distillation, we aim to explore whether it is possible to develop an efficient universal text-to-structure model via instruction tuning on GPT-4o-generated data. To this end, we propose a three-step pipeline for synthesizing text-to-structure data across diverse tasks and schemas. Specifically, we start by identifying a subset of NLP tasks that can be converted into text-to-structure format, and then iteratively expand the pool with new tasks proposed by GPT-4o. Below, we describe the detailed pipeline for constructing the synthetic training data for text-to-structure distillation. Prompt details can be found in \Cref{app:prompt}.

 \paragraph{Task Filtering} While there exists large collections of instruction-tuning data from various sources \cite{wang-etal-2023-self-instruct,wang-etal-2022-super,srivastava2022beyond}, most of the tasks do not fit the definition of a non-trivial text-to-structure task. As we discussed previously, any task output can be trivially converted to JSON format by, for example, rewriting the original output as a JSON object with a single "output" field. Typical examples of such tasks that make up a large portion of available data include classification, translation, and open-ended generation, which usually do not have an inherently structured answer format. Therefore, we aim to identify a set of seed tasks that are suitable for text-to-structure generation. We start from Super-NaturalInstructions \cite{wang-etal-2022-super}, a large collection of more than 1600 NLP tasks where each task in the set comes with a high-level task instruction. Since manually inspecting these tasks can be inefficient, we instead prompt GPT-4o to determine whether a task instruction can be formulated as a meaningful text-to-structure task based on the feasibility of creating an answer schema that satisfies a set of hand-crafted criteria. These criteria impose restrictions on factors such as the number and type of fields, thus excluding schemas such as those with only one field or with meaningless fields that cannot be inferred from the task input itself (e.g., timestamp, metadata, id, etc). The final filtered pool of text-to-structure contains 402 different instructions and their corresponding model-generated answer schema. 

 \paragraph{Task Synthesis} To further diversify and expand the pool of text-to-structure tasks, we prompt GPT-4o to iteratively synthesize new tasks and schemas based on randomly sampled tasks. At each iteration, we provide two pairs of instructions and schemas sampled from the pool, and ask GPT-4o to generate a new pair, subject to the same set of restrictions we impose in the task filtering step.

 \paragraph{Instance Generation and Validation} After we have generated a sufficiently large pool of text-to-structure tasks, we prompt GPT-4o to generate specific task inputs and answers given each pair of instruction and schema. Finally, we validate the generated instance and filter out those with an answer that violates the given schema. The final set of synthesized instruction tuning data consists of $19,819$ instances across diverse tasks and schemas.

\begin{figure}[t]
    \centering
    \includegraphics[width=0.3\textwidth]{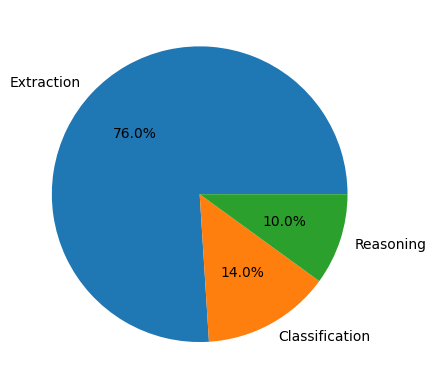}
    \caption{Categories of synthetic text-to-structure data.}
    \label{fig:datapie}
\end{figure}

 \paragraph{Targeted Distillation} In addition to synthesized task instructions, we further augment the instruction tuning data with NER data to improve model capability on entity-related tasks, including NER and RE. We follow \citet{zhou2024universalner} and sample 10K instances from the UniversalNER data constructed from distilling ChatGPT on the Pile corpus \cite{gao2020pile}. We keep the original task inputs but relabel the data with answers generated by GPT-4o.

 To provide a better understanding of our synthetic text-to-structure data, we randomly sample 100 instances from our dataset and manually group the tasks into three categories: classification, extraction, and reasoning. As shown in \Cref{fig:datapie}, a majority of our synthesized data can be categorized as extraction, where the goal is to represent important information from a given piece of text into a structured format. The rest of the data falls either into classification or reasoning tasks. Examples of our synthesized data can be found in \Cref{app:syn}.

 \subsection{Instruction Tuning}
 With a diverse collection of synthetic instruction tuning data, we conduct distillation experiments to study whether we can transfer the strong text-to-structure capabilities of GPT-4o to a much smaller, more cost-efficient model. Specifically, we fine-tune Llama3.1-8B-Instruct model for 1 epoch with a learning rate of 1e-5 and batch size of 64, and then evaluate the fine-tuned model, which we call \bench-8B on our benchmark in the same zero-shot setting.

 As shown in \Cref{tab:sft}, fine-tuning on our synthetic data consistently improves text-to-structure capabilities across different tasks and schemas. Without using any supervised data, \bench-8B outperforms GPT-4o on 5 of 6 NER datasets and RE on CoNLL 2004. \bench-8B also improves performance on datasets like WikiBio, RotoWire, and Meeting Plan, without ever seeing the task schemas during fine-tuning, demonstrating that \bench-8B generalizes well to unseen tasks and schemas. Despite the improvement on challenging tasks such as BFCL and Meeting Plan, there remains a large performance gap largely due to an inherent difference in reasoning and function calling capabilities, and we believe it is unlikely to further close that gap by training on text-to-structure data alone.

\section{Conclusion}
We introduce \bench, a comprehensive benchmark for evaluating text-to-structure generation capabilities of LLMs. By identifying and adapting existing tasks into a unified schema-following text-to-structure problem, \bench provides a holistic view of how well LLMs generate structured answers across diverse tasks and schemas. To facilitate the development of cost-efficient universal text-to-structure models, we propose a pipeline for synthesizing high-quality text-to-structure instruction tuning data with GPT-4o, and demonstrate the possibility of distilling GPT-4o into strong text-to-structure models with significantly fewer parameters.

\section*{Limitations}

In this work, we focus on JSON as the representative structured format due to its versatility and wide use, which simplifies integration with downstream applications such as function call execution and data storage. There also exist many other structured data formats such as XML, LaTeX, HTML, etc. How well modern LLMs support these alternative formats remains an open question, and \bench could potentially be adapted to evaluate different formats as well.

\bibliography{anthology,custom}

\appendix
\onecolumn
\newpage
\section{Prompt Details} \label{app:prompt}

\begin{tcolorbox}[colback=white, colframe=gray!80!gray, boxrule=1pt, title=Task Filtering for Text-to-structure Data Synthesis]

Determine whether the following task requires a structured answer format. If so, say 'YES' and write a json schema to define the answer structure of the task. Otherwise, say 'NO' and explain your reasoning.
\newline

Task: \{task\_instruction\}
\newline

Note:

1. The answer structure of the task should be non trivial and have at least two required fields covering necessary components of the answer. 

2. The answer structure of the task should not include fields that store part or all of the task or the input itself. Only answer fields are allowed.

3. The answer structure of the task should not include metadata (such as timestamp, id) that are not relevant to the task itself.

4. The answer structure of the task should not include fields with ambiguous or subjective values (such as explanation).

5. The json schema should not be tied to any specific input, but applicable to the general task.

6. The json schema should include descriptions of fields.

\end{tcolorbox}

\begin{tcolorbox}[colback=white, colframe=gray!80!gray, boxrule=1pt, title=Task Synthesis for Text-to-structure Data Synthesis]

Propose a new task that requires a structured answer format (start with 'Task: '). Then, write a json schema to define the answer structure of the task (start with 'Schema: '). For example,
\newline

Task: \{task\_instruction\_1\}

Schema: \{task\_schema\_1\}

Task: \{task\_instruction\_2\}

Schema: \{task\_schema\_2\}
\newline

Note:

1. The answer structure of the task should be non trivial and have at least two required fields covering necessary components of the answer.

2. The answer structure of the task should not include fields that store part or all of the task or the input itself. Only answer fields are allowed.

3. The answer structure of the task should not include metadata (such as timestamp, id) that are not relevant to the task itself.

4. The answer structure of the task should not include fields with ambiguous or subjective values (such as explanation).

5. The json schema should not be tied to any specific input, but applicable to the general task. 

6. The json schema should include descriptions of fields.

\end{tcolorbox}

\begin{tcolorbox}[colback=white, colframe=gray!80!gray, boxrule=1pt, title=Instance Generation for Text-to-structure Data Synthesis]

Given the following task instruction and json schema that defines the answer structure, write an example for the task. The example should include an input (start with 'Input: ') that includes all additional information needed to solve the problem, and a json answer (start with 'Answer: ') that strictly follows the json schema.
\newline

Task: \{task\_instruction\}

Schema: \{task\_schema\}

\end{tcolorbox}

\section{\bench Examples} \label{app:bench}
\begin{tcolorbox}[colback=white, colframe=gray!80!gray, boxrule=1pt, title=Named Entity Recognition]

You are a helpful assistant. You must write your answer in JSON format according to the given JSON schema. Do not include anything other than the JSON answer in your response.
\newline

Schema:
\begin{lstlisting}[language=json, numbers=none]
{
  "type": "object",
  "properties": {
    "entities": {
      "type": "array",
      "items": {
        "type": "object",
        "properties": {
          "entity_name": {
            "type": "string"
          },
          "entity_type": {
            "type": "string"
          }
        },
        "required": [
          "entity_name",
          "entity_type"
        ]
      }
    }
  },
  "required": [
    "entities"
  ]
}
\end{lstlisting}

Text: The task is usually to derive the maximum likelihood estimate of the parameters of the HMM given the output sequences.
\newline

Given the above text, your task is to identify and categorize named entities mentioned into the following types: algorithm, conference, product, task, field, metrics, organization, researcher, program language, country, location, person, university.

\end{tcolorbox}

\begin{tcolorbox}[colback=white, colframe=gray!80!gray, boxrule=1pt, title=Text-to-Table Generation]

You are a helpful assistant. You must write your answer in JSON format according to the given JSON schema. Do not include anything other than the JSON answer in your response.
\newline

Schema:
\begin{lstlisting}[language=json, numbers=none]
{
  "type": "object",
  "properties": {
    "party": {
      "type": "string"
    },
    "name": {
      "type": "string"
    },
    "state house": {
      "type": "string"
    },
    "birth date": {
      "type": "string"
    }
  },
  "required": [
    "birth date",
    "name",
    "party",
    "state house"
  ]
}
\end{lstlisting}

Text: gerald kaufman (born june 14, 1932) is a former democratic member of the pennsylvania house of representatives. kaufman, from squirrel hill, served in the house for three terms, sat on committees working on issues of education, industrial development, health and welfare, and chaired the welfare subcommittee of the appropriations committee. during his time in the house, kaufman was known as a "champion of liberal and consumer legislation."
\newline

Given the above text, your task is to extract information according to the schema.

\end{tcolorbox}

\begin{tcolorbox}[colback=white, colframe=gray!80!gray, boxrule=1pt, title=Function Calling]

You are a helpful assistant. You must write your answer in JSON format according to the given JSON schema. Do not include anything other than the JSON answer in your response.
\newline

Schema:
\begin{lstlisting}[language=json, numbers=none]
{
  "type": "object",
  "properties": {
    "base": {
      "type": "integer",
      "description": "The base of the triangle."
    },
    "height": {
      "type": "integer",
      "description": "The height of the triangle."
    },
    "unit": {
      "type": "string",
      "description": "The unit of measure (defaults to 'units' if not specified)"
    }
  },
  "required": [
    "base",
    "height"
  ]
}
\end{lstlisting}

Find the area of a triangle with a base of 10 units and height of 5 units.
\newline

Given the above task, write a function call of the function calculate\_triangle\_area according to the schema.

\end{tcolorbox}

\section{Synthesized Examples} \label{app:syn}
 
\begin{tcolorbox}[colback=white, colframe=gray!80!gray, boxrule=1pt,  title=Synthesized Text-to-structure Data]

Schema:
\begin{lstlisting}[language=json, numbers=none, basicstyle=\tiny\ttfamily]
{
  "title": "Software Release Details",
  "type": "object",
  "properties": {
    "software_releases": {
      "type": "array",
      "description": "A list of software releases from the past year.",
      "items": {
        "type": "object",
        "properties": {
          "software_name": {
            "type": "string",
            "description": "The name of the software."
          },
          "version_number": {
            "type": "string",
            "description": "The version number of the software release."
          },
          "release_date": {
            "type": "string",
            "format": "date",
            "description": "The date when the software was released."
          },
          "new_features": {
            "type": "array",
            "description": "A list of main new features introduced in this version.",
            "items": {
              "type": "string"
            }
          },
          "developer": {
            "type": "string",
            "description": "The developer or company responsible for the software release."
          }
        },
        "required": [
          "software_name",
          "version_number",
          "release_date",
          "new_features",
          "developer"
        ]
      }
    }
  },
  "required": [
    "software_releases"
  ]
}
\end{lstlisting}

In the past year, several software products have been released with new versions. For example, "TechSuite Pro" version 3.5 was released on 2023-05-15 by Tech Innovations Inc. This version introduced features such as enhanced data analytics, improved user interface, and faster processing speeds. Another release was "SecureNet" version 2.1, launched on 2023-08-10 by CyberSafe Solutions. The new version included advanced encryption protocols, multi-factor authentication, and a redesigned dashboard. Additionally, "PhotoEdit Plus" version 4.0 was released on 2023-02-20 by CreativeSoft. This update brought AI-powered editing tools, new filter options, and cloud integration.
\newline

For each software released in the past year, identify and extract key details about the release. This includes the name of the software, the version number, the release date, the main new features introduced in this version, and the developer or company responsible for the release. Return a structured list of software releases with these details.

\end{tcolorbox}

\end{document}